\documentclass[sigconf]{acmart}
\usepackage{adjustbox}
\usepackage{graphics}
\usepackage{amsmath}
\usepackage{enumitem}
\usepackage[ruled,linesnumbered,vlined]{algorithm2e}
\usepackage{algorithmicx}
\usepackage{algpseudocode}
\usepackage{newtxmath}
\usepackage{bbding}
\usepackage{pifont}

\usepackage{multirow}
\usepackage[normalem]{ulem}
\useunder{\uline}{\ul}{}

\usepackage{colortbl}
\usepackage{multicol}
\usepackage{subfigure}
\usepackage{wrapfig}
\usepackage{hyperref}
\usepackage{url}
\usepackage{graphicx}
\usepackage{xspace}
\usepackage{tabulary}
\usepackage{mathtools}
\usepackage{colortbl}
\usepackage{multirow}
\usepackage{booktabs} 
\usepackage{diagbox}
\usepackage{pifont}
\usepackage{hyperref}

\newcommand{\rom}[1]{\uppercase\expandafter{\romannumeral #1\relax}}

\newcommand{\NRdetector}{NRdetector~}
\newcommand{\PUdetector}{NRdetector~}

\AtBeginDocument{%
  }

\copyrightyear{2025}
\acmYear{2025}
\setcopyright{cc}
\setcctype{by}
\acmConference[KDD '25]{Proceedings of the 31st ACM SIGKDD Conference on Knowledge Discovery and Data Mining V.1}{August 3--7, 2025}{Toronto, ON, Canada}
\acmBooktitle{Proceedings of the 31st ACM SIGKDD Conference on Knowledge Discovery and Data Mining V.1 (KDD '25), August 3--7, 2025, Toronto, ON, Canada}
\acmDOI{10.1145/3690624.3709257}
\acmISBN{979-8-4007-1245-6/25/08}

\begin{document}

\title{Noise-Resilient Point-wise Anomaly Detection in Time Series \\Using Weak Segment Labels}

\author{Yaxuan Wang}
\authornote{Both authors contributed equally to this research.}
\affiliation{%
  \institution{University of California, Santa Cruz}
  \streetaddress{1156 High St}
  \city{Santa Cruz}
  \state{CA}
  \country{USA}
  \postcode{95064}
}
\email{ywan1225@ucsc.edu }

\author{Hao Cheng}
\authornotemark[1]
\affiliation{%
  \institution{Hong Kong Baptist University}
  \streetaddress{1156 High St}
  \city{Hong Kong}
  \country{China}
  \postcode{95064}
}
\email{haocheng@comp.hkbu.edu.hk}


\author{Jing Xiong}
\affiliation{%
  \institution{University of California, Santa Cruz}
  \streetaddress{1156 High St}
  \city{Santa Cruz}
  \state{CA}
  \country{USA}
  \postcode{95064}
}
\email{jxiong20@outlook.com}

\author{Qingsong Wen}
\affiliation{%
  \institution{Squirrel AI}
  \city{Bellevue}
  \state{WA}
  \country{USA}
}
\email{qingsongedu@gmail.com}

\author{Han Jia}
\affiliation{%
  \institution{RIPED, CNPC}
  \city{Beijing}
  \country{China}
}
\email{hanjia0217@petrochina.com.cn}
\author{Ruixuan Song}
\affiliation{%
\institution{University of California, Santa Cruz}
  \streetaddress{1156 High St}
  \city{Santa Cruz}
  \state{CA}
  \country{USA}
  \postcode{95064}
}
\email{songrx0704@126.com}
\author{Liyuan Zhang}
\affiliation{%
  \institution{DUT \& RIPED, CNPC}
  \city{Dalian \& Beijing}
  \country{China}
}
\email{zhangliyuan@mail.dlut.edu.cn}
\author{Zhaowei Zhu}
\affiliation{%
  \institution{BIAI, ZJUT \& D5 Data}
  \city{Hangzhou}
  \country{China}
}
\email{zzw@d5data.ai}
\author{Yang Liu}
\authornote{Corresponding author: yangliu@ucsc.edu}
\affiliation{%
  \institution{University of California, Santa Cruz}
  \streetaddress{1156 High St}
  \city{Santa Cruz}
  \state{CA}
  \country{USA}
  \postcode{95064}
}
\email{yangliu@ucsc.edu}

\renewcommand{\shortauthors}{Yaxuan Wang et al.}

\begin{abstract}

Detecting anomalies in temporal data has gained significant attention across various real-world applications, aiming to identify unusual events and mitigate potential hazards. 
In practice, situations often involve a mix of segment-level labels (detected abnormal events with segments of time points) and unlabeled data (undetected events), while the ideal algorithmic outcome should be point-level predictions. Therefore, the huge label information gap between training data and targets makes the task challenging.
In this study, we formulate the above imperfect information as noisy labels and propose NRdetector, a noise-resilient framework that incorporates confidence-based sample selection, robust segment-level learning, and data-centric point-level detection for multivariate time series anomaly detection.
Particularly, to bridge the information gap between noisy segment-level labels and missing point-level labels, we develop a novel loss function that can effectively mitigate the label noise and consider the temporal features. It encourages the smoothness of consecutive points and the separability of points from segments with different labels.
Extensive experiments on real-world multivariate time series datasets with 11 different evaluation metrics demonstrate that NRdetector 
consistently achieves robust results across multiple real-world datasets, outperforming various baselines adapted to operate in our setting. \footnote{ Code is available at \url{https://github.com/UCSC-REAL/NRdetector}.}

\end{abstract}

\ccsdesc[500]{Computing methodologies~Machine learning}
\ccsdesc[300]{Computing methodologies~Anomaly detect}
\ccsdesc[300]{Mathematics of computing~Time series analysis}

\keywords{Time Series Anomaly Detection, Positive and Unlabeled Learning, Learning from Noisy Labels}




\maketitle
\newcommand\kddavailabilityurl{https://doi.org/10.5281/zenodo.14676716}

\ifdefempty{\kddavailabilityurl}{}{
\begingroup\small\noindent\raggedright\textbf{KDD Availability Link:}\\
The source code of this paper has been made publicly available at \url{\kddavailabilityurl}.
\endgroup
}

\section{Introduction}
\label{sec:intro}
Time series anomaly detection (TSAD) plays a critical role in many real-world monitoring systems and applications, such as robot-assisted systems~\cite{park2018multimodal}, space exploration~\cite{hundman2018detecting}, and cloud computing~\cite{zhang2021cloudrca}. It is the task of discerning unusual or anomalous samples in time series data. Detecting highly anomalous situations is essential for avoiding potential risks and financial losses. Due to the complex temporal dependencies and limited label data, the most dominant research focuses on building the normal patterns from the time series data under an unsupervised setting and can be generalized to unseen anomalies~\cite{park2018multimodal,li2022learning,shen2020timeseries,zhang2022time,su2019robust,xu2021anomaly,chen2023lara,xu2024calibrated}. However, the performance of these unsupervised learning methods is constrained by the lack of prior knowledge concerning true anomalies~\cite{elaziz2023deep}. They are not good at finding specific anomalous patterns, especially when the anomalies are embedded within the training data for building the normal patterns. 
An intuitive approach to training a model is through supervised learning, using point-level labels for guidance.  However, labeling every anomalous time point is neither practical nor precise due to the significant time and cost required for accurate identification. 

Several studies~\cite{lee2021weakly,liu2024treemil} focus on distinguishing anomalous time points from normal ones using segment-level labels for model training. Acquiring weak labels by simply indicating the occurrence of anomalous events is a more practical approach for real-world applications. In the segment-level setup, consecutive time points of a specified length are grouped into a segment, and the labels are provided only for the segment level rather than for individual time points~\cite{zhou2004multi,perini2023learning}.
Specifically, a \textit{positive segment label} indicates that at least one time point within the segment is anomalous, while \textit{a negative segment label} restricts all the points in the segment are normal. One limitation of this setting is that it may overlook potential label errors commonly found in real-world scenarios. In real-world TSAD problems, a positive label can be seen as a true annotation because an observed and recorded abnormal event is often verified. However, the other events may be either normal or abnormal events since the abnormal behavior may be missed. 
Therefore, detecting point-level anomalies given only segment-level labels with label errors is a practical approach for real-world applications. In this setting, previous methods~\cite{lee2021weakly,liu2024treemil}  fail to effectively address the noise within segment-level labels, particularly when abnormal behavior occurs in unverified events.

To overcome the challenges of 1) label errors in segment-level labels and 2) the lack of point-wise labels in TSAD, we propose a two-stage solution:\\
\noindent$\bullet$ Stage-1: Coarse-grained PU learning: The coarse-grained segment-level labels are formulated by Positive and Unlabeled (PU) learning~\cite{ma2017pu, de2022network,luo2021pulns} problem, i.e., only positive segment labels are provided along with unlabeled segments.\\
\noindent$\bullet$ Stage-2: Fine-grained abnormal point detection: We use a data-centric method to score which point is more likely to be abnormal and automatically calculate the best threshold to filter out the abnormal points from a positive segment. 

\begin{figure}[!t]
\centering
\includegraphics[width=0.44\textwidth]{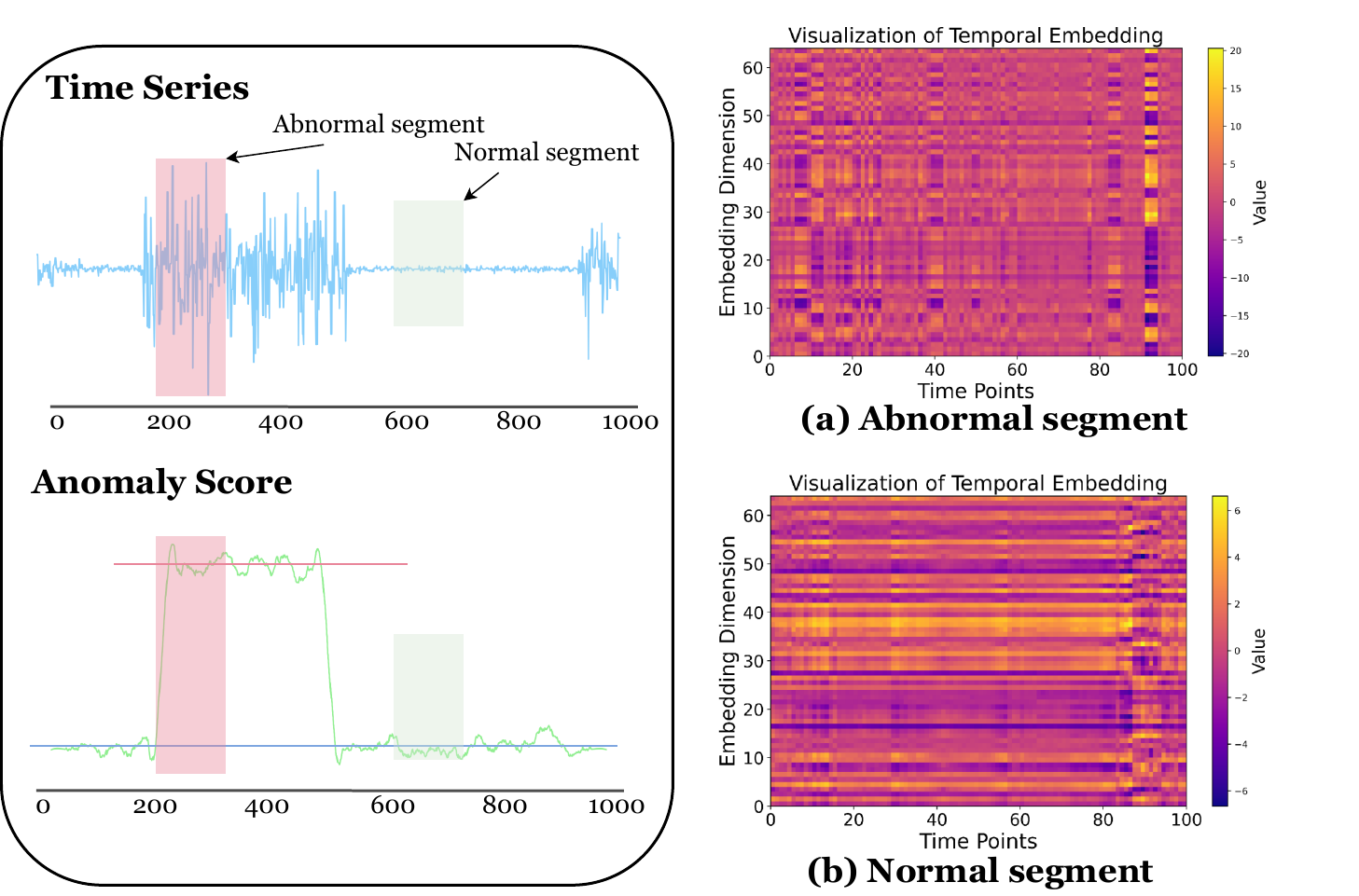}
\caption{Illustration of the insights. The x-axis in (a) and (b) represents the time step, and the y-axis represents the feature vector. When true point labels are accessible, we observe: a) \textbf{Continuity:} The point-level anomaly scores vary smoothly between points within either abnormal (red) or normal (green) segments; 
b) \textbf{Discriminability:} The range of scores (as in the color bar) of anomalous segments ($\pm$20) is sufficiently different from the normal segments ($\pm$6).
When point labels are missing, the learned embeddings should also preserve both properties.
} 
\vspace{-3mm}
\label{fig:insight}
\end{figure}

There is a significant information gap between the noisy segment-level labels and the true point-level labels. To bridge this gap when the true point-level labels are missing, we propose a novel loss function that leverages the properties of point-level temporal embeddings to regularize the learning process. Our loss function is motivated by Figure~\ref{fig:insight}, which shows a particular time-series anomaly detection task with \textbf{\textit{true}} point-level labels. The left subfigure shows that abnormal segments are visually distinguishable based on their values in the time series. Moreover, when the true point labels are available, this discriminability is also evident in their temporal embeddings after training, as seen in the range of embedding values in Figure~\ref{fig:insight}(a) ($\pm$20) and Figure~\ref{fig:insight}(b) ($\pm$6). In the meanwhile, there is continuity between consecutive time points in terms of their embeddings in both normal and abnormal segments. The observation motivates us to design a novel loss function in a contrastive fashion, i.e., during training, when the point labels are \textbf{\textit{missing}}, the ideal embedding should also preserve both properties. Therefore, we regularize the temporal embedding to encourage the discriminability between points in segments with different (noisy) labels and the continuity between consecutive points.

Specifically, we introduce NRdetector, a noise-resilient multivariate TSAD framework based on PU learning under inaccurate and inadequate segment-level labels. Our framework includes a coarse-grained PU learning stage followed by a fine-grained abnormal point detection stage. 
In Stage-1, we first reduce the label noise rate of the unlabeled set by a confidence-based sample selection module, then apply our designed loss function to guide the segment-level learning and make the output aligned with the time points.
In Stage-2, we propose a data-centric point-level detection approach to further determine the point-level anomalies, where the ratio is determined by a training-free automated estimator~\cite{zhu2023unmasking,zhu2021clusterability}.
Finally, we can get the prediction of all time points. 
The contributions of this paper are summarized as follows.

\begin{itemize}[leftmargin=*]
\item We focus on a novel and practical scenario in TSAD, where abnormal labels are limited and coarse-grained, indicating a time range rather than an exact time point due to challenges like labeling ambiguity or imprecise event timing. To address these issues, we reformulate the problem as a PU learning task and propose NRdetector, a noise-resilient framework designed to handle the imperfect information inherent in this setting.
\item To handle the information gap between noisy segment-level labels and missing point-level labels, we propose a new loss function in a contrastive manner, encouraging the smoothness of consecutive points and the separability of points from segments with different labels.
\item Experiments on five real-world multivariate time series benchmarks demonstrate that \NRdetector consistently surpasses existing TSAD methods in noisy labels setting. We also perform comprehensive ablation studies to evaluate the proposed method.
\end{itemize}

\section{Related Work}
\textbf{Time Series Anomaly Detection.} Extensive research has explored TSAD using deep neural networks in unsupervised settings~\cite{park2018multimodal,ruff2018deep,shen2020timeseries,zhang2022tfad,xu2021anomaly,xu2018unsupervised, wu2022timesnet,zhou2023one,sun2023unraveling,kim2023model,lai2024nominality}. These methods aim to learn models that accurately describe normality and identify anomalies as deviations from these patterns~\cite{yang2023dcdetector}. However, the lack of anomaly labels during training makes it challenging to detect different anomalies accurately. Several studies utilize full supervised learning to detect the anomaly time points~\cite{gao2020robusttad,ryzhikov2021nfad,carmona2021neural}, but obtaining comprehensive point-level labels is impractical due to the labor-intensive nature of data collection. Additionally, label imbalance can hinder supervised model development. Augmentation-based methods~\cite{darban2025carla,wang2024cutaddpaste, carmona2021neural} can partially address this issue by expanding known anomaly data. Our approach differs from augmentation-based methods by leveraging real, coarse-grained anomaly labels to address a practical setting, whereas those methods rely on synthetic anomalies and typically use window-based slicing, making point-level anomaly determination within each sample unclear.
Weakly supervised approaches~\cite{lee2021weakly,liu2024treemil,sultani2018real} optimize models to classify segments accurately by leveraging segment-level labels. 
In this paper, we focuses on detecting fine-grained point-level anomalies using partial positive (anomalous) segment-level labels, which allows us to effectively handle real-world scenarios with noisy label information.

\textbf{Learning with Noisy Labels.} In the literature on learning with label noise, popular approaches can be divided into several categories, such as robust loss design~\cite{liu2020peer,zhu2021second}, transition matrix estimation~\cite{patrini2017making,zhu2021clusterability}, and sample selection~\cite{han2018co}. Among all these methods, the most popular work concentrates on developing risk-consistent strategies, i.e., conducting empirical risk minimization (ERM) with custom-designed loss functions on noisy distributions results in the same minimizer as conducting ERM on the corresponding unseen clean distribution. We aim to train a classifier using a small number of labeled positive examples alongside a larger quantity of unlabeled ones, where the unlabeled data is considered to have noisy labels.

\textbf{Positive and Unlabeled learning.} Positive and Unlabeled (PU) Learning is a special form of semi-supervised learning (SSL). 
Compared with traditional SSL, this task is much more challenging due to the absence of any known negative labels. Existing PU learning methods typically fall into two categories. The first category involves sample-selection-based approaches, which involve negative sampling from unlabeled data using either handcrafted heuristics or standard SSL techniques \cite{liu2002partially, li2003learning, luo2021pulns, de2022network, ma2017pu, nguyen2011positive}. The second category employs a cost-sensitive approach, treating all unlabeled segments as potentially negative and adjusting the objective's estimation bias through carefully designed misclassification risks~\cite{du2014analysis,kiryo2017positive}. The mainstream of PU methods has shifted towards the framework of cost-sensitive learning~\cite{chen2020self,hu2021predictive,zhao2022dist,zhu2023robust}. PU Learning has been employed in the field of anomaly detection~\cite{zhang2017positive, perini2023learning}, including time series~\cite{nguyen2011positive, zhang2021robust} and video anomlay detection~\cite{elaziz2023deep,zhang2023exploiting}.
Our work focuses on segment-level PU learning, where we have partial knowledge of the weak labels for abnormal temporal segments. We consider both sample-selection-based and cost-sensitive approaches to propose a more robust and noise-resilient TSAD detector.

\begin{figure*}[!hbtp]
\centering
\includegraphics[width=0.9\textwidth]{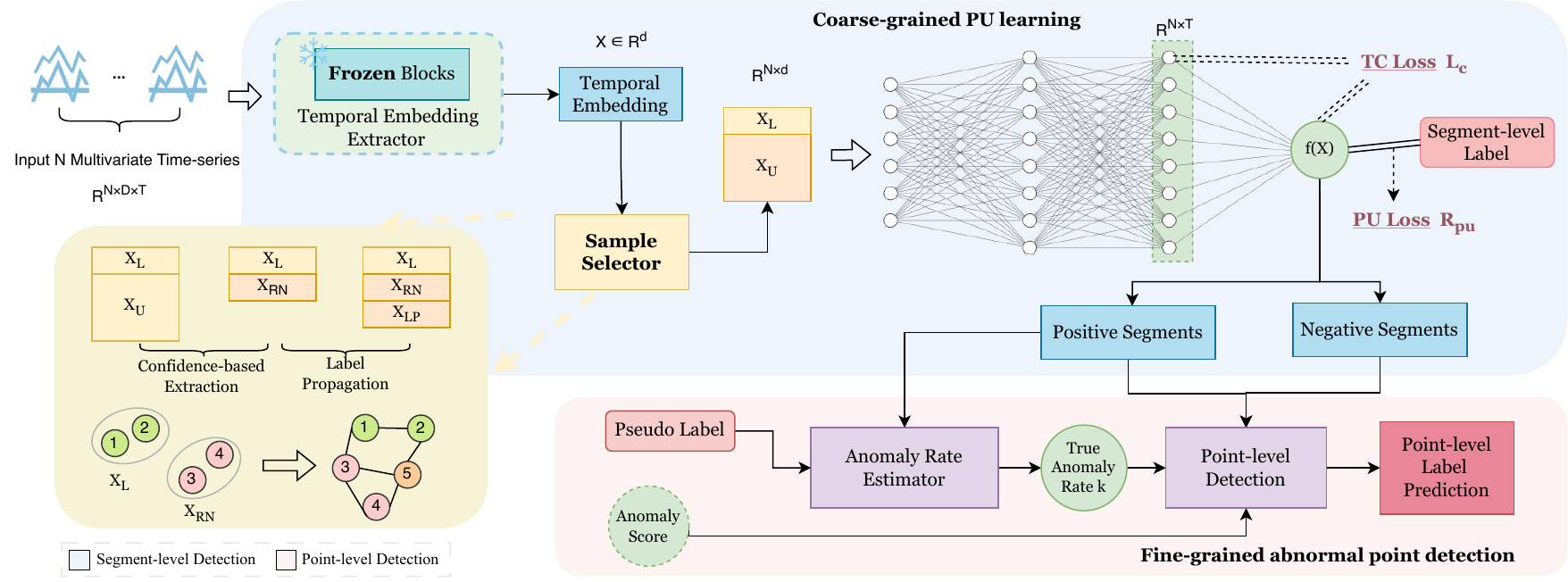}
\caption{
The workflow of the \PUdetector framework. \PUdetector consists of two main stages: coarse-grained PU learning and fine-grained abnormal point detection. In the Sample Selector module, $\mathcal X_L$ denotes the set of labeled positive segments, $\mathcal X_U$ denotes the set of unlabeled segments, $\mathcal X_{RN}$ denotes the set of extracted reliable negatives based on the confidence scores, and $\mathcal X_{LP}$ denotes the set of likely negatives after the label propagation process. $f(X)$ represents the output from the last linear layer of our classifier, processed through a Sigmoid function. TC Loss is the time constraint loss and PU Loss is the non-negative PU risk estimator in Section~\ref{sec:criterion}.
}
\label{fig:framework}
\end{figure*}

\section{Preliminaries}

We summarize the key concepts and notations as follows.

\subsection{Problem Formulation}
We consider a time series anomaly detection problem that aims to identify \textit{anomalous time points} from the \textit{input segments} of temporal data, where each segment can be collected for a fixed period of time $T$ or obtained by splitting time series into fixed-length temporal data~\cite{lee2021weakly}. Formally, given a $D$-dimensional \textit{input segment} $ \hat{X}=\left[ \hat{x}^1,\hat{x}^2,...,\hat{x}^T \right] \ \in \ \mathbb{R}^{D\times T}$ of temporal length $T$, we aim to obtain the point-level anomaly predictions $\hat{y}\in \left\{ 0,1 \right\}$ for each $\hat x^t$, where with $\hat{y} = 0$ for negative (normal) points and $\hat{y} = 1$ for positive (abnormal) points. Let $X\in \mathbb{R}^d$ be the representation vector of the segment $\hat{X}$, containing the temporal features with $d$ dimensions. Let $Y\in \left\{ 0,1 \right\}$ be the segment-level labels (also as weak labels), indicating if there is at least one anomalous point within the segment. By decoupling the extraction of the time dependency, we could consider the problem as an extremely unbalanced binary classification task (in terms of segment-level predictions) with only partial segment-level positive labels. 
For detecting the anomaly time points, we need to develop both segment and point-level methods to assign labels to the segments and time points.

\subsection{Learning with Noisy Labels}
The traditional segment-level TSAD task with clean labels~\cite{wei2021learning} often builds on a set of N training segments denoted by $D:=\left\{ \left( X_n,Y_n \right) \right\} _{n\in \left[ N \right]}$, where $\left[ N \right] :=\left\{ 1,2,...,N \right\} $, $X_n$ is the n-th segment embedding, and $Y_n$ is the label of the n-th segment, which indicates whether anomalous events are observed in the segment. 
In our considered segment-level classification problem, instead of having access to the clean dataset D, the trained classifier could only obtain a noisy dataset $\tilde{D}:=\{ ( X_n,\tilde{Y}_n ) \} _{n\in [ N ]}$, where the noisy label $\tilde{Y}_n$ may or may not be the same as $Y_n$. 
Statistically, the random variable of noisy labels $\tilde{Y}$ can be characterized by transition probabilities~\cite{patrini2017making,wei2021smooth, zhu2021clusterability}, i.e., $e_0:=P( \tilde{Y}=1|Y=0 ) $, $e_1:=P( \tilde{Y}=0|Y=1 )$, representing the probability of flipping the clean label $Y=i$ to the noisy label $\tilde{Y}=j$.

\subsection{Positive and Unlabeled Learning}
Let $p\left( x,y \right)$ be the underlying probability density, and $p\left(x\right)$ be the marginal distribution of the input. Then the sets of actual positive and negative segments could be denoted as: $\mathcal X_P=\left\{ X \right\} ^{N_P} \sim p_P\left( x \right)$, and $\mathcal X_N=\left\{ X \right\} ^{N_N} \sim p_N\left( x \right) $, where $p_P\left( x \right)$ and $p_N\left( x \right)$ denote the class-conditional distribution of positive and negative segments respectively. The whole training input set $\mathcal X= \mathcal X_P\cup \mathcal X_N=\left\{ X \right\} ^N \sim p\left( x \right) $ and the marginal distribution can be formulated as $p\left( x \right) =\pi _P\cdot p_P\left( x \right) +\pi _N\cdot p_N\left( x \right)$, where $N=N_P+N_N$, $\pi _P=P\left( Y=1 \right)$ is the class prior probability and $\pi _N=1-\pi _P$. 

PU learning is a special case, in the way that only a small portion of positive data are labeled. Formally, the input of training set $\mathcal X_{PU}=\mathcal X_L\cup \mathcal X_U$ where $\mathcal X_L$ or $\mathcal X_U$ represents the labeled positive or the unlabeled subset of both negative and positive segments, respectively. 
In our paper, we consider the scenario~\cite{bekker2020learning} which is practical that positives are labeled uniformly at random and independently of their features, while the unlabeled data are, i.i.d drawn from the real marginal distribution: $\mathcal X_L=\left\{ X \right\} ^{N_L} \sim p_P\left( x \right) $, $\mathcal X_U=\left\{ X\right\} ^{N_U} \sim p_P\left( x \right) =p\left( x \right) $, where $N=N_L+N_U = N_P + N_N$. Note that the labeled segment set is much smaller than the unlabeled set, that is $N_L\ll N_U$.

The classification task aims to identify a classifier $f\in F:\mathbb{R}^d\rightarrow \mathbb{R}$ that outputs the anomaly score of the segment, where 
$f\left(X\right)\in\left[0,1\right]$. A hyperparameter threshold $\tau$ is used to decide if the segment is anomalous ($f(X)>\tau$) or not. Also known as the expected empirical risk:
\begin{equation}
R=\mathbb{E}_{\left( X,Y \right) \sim p\left( x,y \right)}\left[\vmathbb{1}( \hat{Y} ,Y) \right], 
\label{eq:OriRisk}
\end{equation}
with $\hat{Y} \in \left\{ 0,1 \right\}$ denoting the predicted label of segment $X$ by $f$ based on the threshold; $\vmathbb{1}(\hat{Y} ,Y )$ refers to the 0-1 loss, which equals 0 when $\hat{Y}=Y$ and otherwise 1. 
One common choice is training a deep neural network (DNN) by minimizing the empirical risk: $\hat{f}=arg\min_f\frac{1}{N}\sum_{n=1}^N{\mathcal{L}\left( f\left( X_n \right) ,Y_n \right)}$. Notation $\mathcal{L}\left( \cdot \right) $ represents the surrogate loss function~\cite{bartlett2006convexity}, i.e., cross-entropy loss.

We treat the corresponding dataset of input $\mathcal X_{PU}$ as the noisy dataset $\tilde{D}$ of the clean dataset $D$, thus we can get a special form of the noise transition matrix, where $e_0$ is equal to 0 in the PU setting. Note that we don't have access to the point-level labels.

\section{Methodology}
In this section, we introduce a noise-resilient time series anomaly detection approach designed to identify point-level anomalies by leveraging the weak segment labels. Figure~\ref{fig:framework} illustrates the overall framework of the NRdetector.
\subsection{Overall Framework}
Our framework includes Stage-1 (confidence-based sample selection and robust segment-level learning) and Stage-2 (data-centric point-level detection). At the initial stage, we extract the Temporal Embedding of the input multivariate time series (Section~\ref{seg:embed}). The extracted temporal dependencies will be learned through the pre-trained informative representation model. 
In the sample selection module (Section~\ref{sec:selector}), we employ the Sample Selector based on the confidence score to identify reliable and likely negative segments before training, aiming to reduce the imbalance between $\mathcal X_L$ and $\mathcal X_U$ and lower the proportion of corrupted segment-level labels (actual positive labels) in the training dataset.

In the robust segment-level learning module (Section~\ref{sec:criterion}), the \PUdetector employs a six-layer MLP classifier equipped with a specially designed loss function to guide the learning process. In the training process, we just see all selected unlabeled segments as negative (normal), which might result in a negative prediction preference of the classifier. Thus, the PU Criterion module plays a crucial role in our design. It can correct the problem of overfiting the negative segments and consider the temporal feature within the segments. 

More specifically, the PU criterion comprising two components: (1) the PU loss, designed to effectively mitigate noise in the segment-level labels and accurately classify input segments according to their anomaly labels; and (2) the Time Constraint (TC) loss, which bridges the gap between noisy segment-level labels and missing point-level labels and also considers temporal features within segments.
Note that the temporal segments here are actually a sequence of temporal representations of consecutive time points. Thus, we can ignore the different time series outliers~\cite{lai2021revisiting}, like the point-wise outliers and pattern-wise outliers. Based on this, we know that normal segments tend to share the same latent pattern, while anomalous segments may have different patterns, but there's one stable thing: the pattern inside the segment itself should be consistent. The key insight is that the point-level anomaly scores should vary smoothly between points within one segment. And the average scores of the anomalous segments ought to be greater than those of the normal segments. We can distinguish anomalous segments from normal segments with the well-designed PU Criterion. 

After obtaining the segment-level prediction, we propose a data-centric point-level detection approach to further determine the point-level anomalies (Section~\ref{sec:point}). 

\subsection{Stage-1: Coarse-Grained PU Learning}

\subsubsection{Temporal Embedding}
\label{seg:embed}
This module projects the time-series data into the required feature dimensions.
To extract temporal features, we utilize the basic architecture of dilated CNN (DiCNN) as introduced by WaveNet~\cite{oord2016wavenet}.
We regard the current unlabeled data as negative and the abnormal data in the labeled data as positive and put it into the WETAS~\cite{lee2021weakly} framework. Note that the extractor here can be replaced with another temporal feature extractor~\cite{zhou2023one,yang2023dcdetector}. Then, we interpret the model output vector $h_t$ at each point to be a temporal representation of that point. The segment-level temporal embedding is obtained by global average pooling (GAP)~\cite{lin2013network} on the output vectors along the time axis. 
Additionally, we determine the anomaly score $S$ of each time point through the use of the anomaly weight $w \in \mathbb{R}^{d}$ and sigmoid function $\sigma$.
\begin{equation}
\begin{aligned}
     X=\text{GAP}(h_1,h_2,...,h_T)\ \in \mathbb{R}^d,  \qquad S=\sigma ( w^TX ).   
\end{aligned}
\label{eq:embed}
\end{equation}

\subsubsection{Sample Selector}
\label{sec:selector}
To enlarge the differences between the anomalous and normal patterns and reduce the imbalance in the training dataset, we propose a network-based unlabeled negative segments selector based on the confidence scores. This selector identifies the most reliable negative segments within the unlabeled dataset $\mathcal X_U$, which consists of noisy negatives, to create a refined unlabeled dataset but with fewer likely anomalous data. 
We calculate the confidence scores of each segments in $\mathcal X_U$ using the cosine similarity metric. Segments that are most dissimilar to the segments in $\mathcal X_L$ will have higher confidence scores. Based on the scores, we formulate a reliable negative segments set $\mathcal X_{RN}$.
Following the assumptions presented in~\cite{de2022network}, we use Katz index~\cite{ma2017pu} to measure the similarity between each segment $X$.
We also created a KNN network, where the segments are represented as nodes. Based on the ideas that the class information of neighboring segments should be similar and the class information of labeled segments assigned during the process must be consistent with known label information (amomalous), we utilize label propagation~\cite{de2022network} to minimize the function and obtain the corresponding propagated label $F$:
\begin{equation}
\begin{aligned}
L\left( F \right) \ =\ \frac{1}{2}\sum_{X_i,X_j\in \mathcal X_L\cup \mathcal X_U}{w_{X_i,X_j}\varOmega ( F_{X_i},F_{X_j}) +\mu \sum_{X_i\in \mathcal X_L}{\varPhi ( F_{X_i},\bar{Y}_{X_i})}},
\end{aligned}
\label{eq:lpo}
\end{equation}
where $\varOmega$ and $\varPhi$ are distance functions, $F$ is the class information vector, $F \in \mathbb{R}^{N\times 2}$; $F_{X_i},F_{X_j}\in \mathbb R^2$, indicating the propagated label information of certain segments $X_i$ and $X_j$; $w_{X_i,X_j}$ is the similarity between $X_i$ and $X_j$; $\Bar{Y}_{X_i}$ is the given label of positive segments, denoting the vector $[0,1]$. The detailed algorithms are in Appendix~\ref{app:algorithms}. By using this algorithm, we can create a new unlabeled dataset $\bar{\mathcal X}_U=\mathcal X_{RN}\cup \mathcal X_{LP}$ based on $F$, where $\mathcal X_{LP}$ refers to those reliable negatives by the label propagation process.

Intuitively, implementing the selector can reduce the label noise rate while at the cost of losing some training samples. When the benefit of label noise rate reduction is more significant than the harm of losing training samples, the performance can be improved. We defer more analyses to Appendix~\ref{app:theo-bound}.

\subsubsection{PU Criterion}
\label{sec:criterion}
In the robust segment-level learning module, we  formalize a loss function based on the Non-negative Risk Estimator~\cite{kiryo2017positive} and the designed Time-Constraint Loss, which considers the underlying temporal structure of the anomalous and normal segments.
The last layer of our classifier is the anomaly score $f(X)$ of the segment and the previous layer is the T-dimension output $h(X)$, which can be seen as the anomaly scores of each point within the T-length segment.

\textbf{Non-negative PU Risk Estimator (PU Loss).} In traditional binary classification task, thanks to the availability of $\mathcal X_P$ and $\mathcal X_N$, $R\left(f\right)$ can be approximated directly by:
\begin{equation}
    \begin{aligned}
        R\left( f \right) =\pi _P \cdot \mathbb E_{X\sim p_P\left( x \right)}\left[\ell( f\left( X \right) ,1 ) \right] +\pi _N \cdot \mathbb E_{X\sim p_N\left( x \right)}\left[ \ell\left( f( X ) ,0 \right) \right] 
    \end{aligned}
    \label{eq:r1}
\end{equation}
where $\ell$ is by default the 0-1 loss $\ell_{01}$; when used for training, $\ell_{01}$ is replaced with a surrogate loss. Based on distribution alignment~\cite{zhao2022dist}, i.e., $\mathbb E_{\left( X,Y \right) ~p_P\left( x,y \right)}Y=\pi _P$, we can formulate the optimization objective as follows:
\begin{equation}
    \begin{aligned}
        R_{pu}=2\pi_P \left| \frac{1}{N_L} \sum_{X\in \mathcal X_L}{f(X)} -1 \right|+\left|\frac{1}{N_U}\sum_{X\in \bar{\mathcal X}_U}{f(X)} -\pi_P\right|
    \end{aligned}
    \label{eq:r2}
\end{equation}
Though the PU Loss $R_{pu}$ is biased, Proposition 1 in ~\cite{zhao2022dist} shows that the original risk can be upper bounded by $R_{pu}$, so optimizing the upper bound naturally leads to a minimization of the original risk.

\textbf{Time-Constraint Contrastive Loss.} One limitation of the PU Loss function is that it ignores the underlying temporal structure within the data, although the features already contain the temporal dependency. Motivated by Figure~\ref{fig:insight}, we aim to make the loss function handle the information gap between noisy segment-level labels and missing point-level labels. Since the segment is a sequence of time points, the anomaly score should vary smoothly within segments. Therefore, we enforce temporal smoothness between anomaly scores of temporally adjacent points within the same segments by minimizing the difference of scores for adjacent time points as:
\begin{equation}
\begin{aligned}
L_{\text{smooth}}=\frac{1}{N}\sum_{i=1}^N{\sum_{j=1}^{T-1}{\left( {h(X_{i})}^{j}  -{h(X_{i})}^{j+1}  \right)^2}},
\end{aligned}
\label{eq:smooth}
\end{equation}
where $h\left( \cdot \right) $ denotes the corresponding anomaly score for each point. Also, we want the anomalous segments to have higher anomaly scores than the normal segments. Thus, we have the separability term:
\begin{equation}
\begin{aligned}
L_{\text{sep}}=\frac{1}{N_U}\sum_{X\in \bar{\mathcal X}_U}{f\left( X \right)}-\frac{1}{N_L}\sum_{X\in \mathcal X_L}{f\left( X \right)}
\end{aligned}
\label{eq:sepa}
\end{equation}
By incorporating the smoothness and separability constraints on the segment and point scores, we design a Time-Constraint loss function (TC Loss):
\begin{equation}
\begin{aligned}
 L_{c}=\lambda _1\cdot L_{\text{smooth}}+\lambda _2 \cdot L_{\text{sep}},
\end{aligned}
\end{equation}
where $\lambda_1$ and $\lambda_2$ control the strength of $L_{\text{smooth}}$ and $L_{\text{sep}}$, respectively. Note that we use batch-wise implementation. 

Finally, we present the overall objective of our PU Criterion with PU Loss and time constraint loss, denoted as follows:
\begin{equation}
\begin{aligned}
L=R_{pu}+\lambda \cdot L_{c}  
\end{aligned}
\label{eq:overall}
\end{equation}
where $\lambda$ adjusts the importance of $L_{c}$.

\subsection{Stage-2: Fine-Grained Point Detection}
\label{sec:point}
After obtaining the predicted positive segments, we need to further detect the anomalous points. In the data-centric point-level detection module, the points are considered negative in predicted normal segments. For all positive segments, we rank all points according to the anomaly score of each point and then take a user-specified anomaly rate k, where k is a hyperparameter, to get pseudo-predicted labels. For example, when k=0.5, the top 50\% of the sorted points are regarded as pseudo anomaly points. To get the true anomaly rate in these time points, we put the pseudo prediction result and the feature vector (or simply anomaly scores $S$) corresponding to each point from the temporal embedding model into the anomaly rate estimator~\cite{zhu2021clusterability} and get the probability of the clean label to automate the threshold selection process.
The predicted true anomaly rate is then used to select abnormal time points from all sorted points. In this way, we get the point-level predictions of the time series. 

\section{Experiments}
\subsection{Benchmark Datasets}
\textbf{Dataset.} For our experiments, we use five temporal datasets collected from various tasks to detect anomalous points. (1) \textbf{Electromyography Dataset (EMG)~\cite{lobov2018latent}} records 8-channel myographic signals via bracelets worn on subjects' forearms.
(2) \textbf{Server Machine Dataset7 (SMD)~\cite{su2019robust}} records the 38 dimensions multivariate time series from Internet server machines over five weeks. 
(3) \textbf{Pooled Server Metrics dataset (PSM)~\cite{abdulaal2021practical}} records 25 dimensions data from multiple eBay server machines.
(4) \textbf{Mars Science Laboratory (MSL)~\cite{hundman2018detecting}} records 55 dimensions data from Mars rover.
(5) \textbf{Soil Moisture Active Passive (SMAP)~\cite{hundman2018detecting}} records 25 dimensions data and presents the soil samples and telemetry information.
The details of the five benchmark datasets are in Table~\ref{tab:data}. 
\begin{table*}
\caption{Details of five benchmark datasets. TPS represents the number of the positive segments within the training data. AR (anomaly ratio) represents the abnormal time points proportion of the whole dataset.}
\vspace{-3mm}
\begin{tabular}{c|cccccccc}
\hline \hline
 Benchmark & Time Points  & Source &\#Training &\#Testing & \#TPS & Dimensions & AR (\%)  \\ \hline 
EMG & 423825 & Myographic Signal Machine & 304400 & 130900   & 222 & 8  & 5.8 \\
SMD & 708420 & Internet Server Machine & 495870 & 212550   & 463 & 38  & 4.2 \\
PSM & 87841 & eBay Server Machine & 61488 & 26353   & 191 & 25  & 27.8 \\
MSL & 73900 & NASA Space Sensors & 51610 & 22119   & 99 & 55  & 10.5 \\
SMAP & 427617 & NASA Space Sensors & 299331 & 128286   & 506 & 25  & 12.8 \\
\hline \hline
\end{tabular}
\label{tab:data}
\end{table*}

We split the set of all segments by 7:3 ratio into training and test sets. 
The average anomalous ratio (AR) of these datasets is 12.22\% (point-level).
We construct the positive dataset for training by providing only 40\% of the anomalous segment-level labels, resulting in a label noise rate of 0.6 (segment-level). The labels for the remaining segments (both anomalous and normal) are not given to construct the unlabeled dataset. We have no access to the point-level labels. Each segment consists of $L=100$ continuous time points. 

\begin{table*}[]
\caption{Performances of \PUdetector and other baselines on real-world multivariate datasets. $F1$ is the F1-score, $F1_{PA\%K}$ is the optimized PA-based F1 score. The best ones are in \textbf{Bold}, and the second ones are \underline{underlined}.}
\vspace{-3mm}
\centering
\begin{tabular}{c|cc|cc|cc|cc|cc}
\hline\hline
\textbf{Dataset} & \multicolumn{2}{c|}{\textbf{EMG}} & \multicolumn{2}{c|}{\textbf{SMD}} & \multicolumn{2}{c|}{\textbf{PSM}} & \multicolumn{2}{c|}{\textbf{MSL}} & \multicolumn{2}{c}{\textbf{SMAP}} \\ 
\hline
\textbf{Metric} & \multicolumn{1}{c}{\textbf{$F1$}} & \multicolumn{1}{c|}{\textbf{$F1_{PA\%K}$}} & \multicolumn{1}{c}{\textbf{$F1$}} & \multicolumn{1}{c|}{\textbf{$F1_{PA\%K}$}} & \multicolumn{1}{c}{\textbf{$F1$}} & \multicolumn{1}{c|}{\textbf{$F1_{PA\%K}$}} & \multicolumn{1}{c}{\textbf{$F1$}} & \multicolumn{1}{c|}{\textbf{$F1_{PA\%K}$}} & \multicolumn{1}{c}{\textbf{$F1$}} & \multicolumn{1}{c}{\textbf{$F1_{PA\%K}$}} \\ 
\hline
DCdetector & 0.0155 & 0.0259 & 0.0105 & 0.0178 & 0.0169 & 0.0272 & 0.0196 & 0.0323 & 0.0169 & 0.0272  \\
Anomaly Transformer & 0.0116 & 0.0196 & 0.0181 & 0.0296 & 0.0195 & 0.0334 & 0.0161 & 0.0261 & 0.0173 & 0.0257\\
AutoFormer & 0.0070 & 0.0128 & 0.0646 & 0.1060 & 0.0532 & 0.0821 & 0.0639 & 0.1086 & 0.0091 & 0.0163 \\
FEDformer & 0.0274 & 0.0484 & 0.0934 & 0.1500 & 0.0494 & 0.0787 & 0.0641 & 0.1090 & 0.0156 & 0.0283 \\
TimesNet & 0.0089 & 0.0168 & 0.0959 & 0.1373 & 0.0172 & 0.0306 & 0.0731 & 0.1210 & 0.0126 & 0.0216 \\
One-fits-all & 0.0388 & 0.0668 & 0.0912 & 0.1381 & 0.0253 & 0.0425 & 0.1234 & 0.1848 & 0.0101 & 0.0172 \\ 
\hline
AutoFormer++ & 0.0074 & 0.0134 & 0.0679 & 0.1106 & 0.0855 & 0.1319 & 0.0637 & 0.1082 & 0.0087 & 0.0160 \\ 
FEDformer++ & 0.0299 & 0.0527 & 0.0960 & \underline{0.1535} & 0.0892 & 0.1371 & 0.0637 & 0.1083 & 0.0152 & 0.0279 \\ 
TimesNet++ & 0.0102 & 0.0186 & 0.0984 & 0.1412 & 0.0342 & 0.0525 & 0.0694 & 0.1154 & 0.0135 & 0.0230 \\ 
One-fits-all++ & 0.0420 & 0.0709 &  0.0884 & 0.1335 &  0.0331 & 0.0535 & \underline{0.1326} & \underline{0.1939} & 0.0139 & 0.0177 \\ 
\hline
DeepMIL & 0.0548 & 0.0936 & 0.1009 & 0.1319 & 0.2251 & 0.3279 & 0.0103 & 0.0178 & 0.1024 & 0.1318 \\ 
WETAS & \underline{0.2446} & \underline{0.3246} & \underline{0.1020} & 0.1171 & \underline{0.3927} & \underline{0.4891} & 0.1113 & 0.1714 & 0.1559 & 0.1775 \\ 
TreeMIL & 0.1696 & 0.2440 & 0.0999 & 0.1164 & 0.3494 & 0.4208 & 0.1258 & 0.1474 & \underline{0.2079} & \underline{0.2397} \\ 
\hline
\textbf{\PUdetector} (Ours) & \textbf{0.4431} & \textbf{0.5174} & \textbf{0.1092} & \textbf{0.1630} & \textbf{0.4990} & \textbf{0.6356} & \textbf{0.2029} & \textbf{0.2219} & \textbf{0.2367} & \textbf{0.2906} \\ 
\hline\hline
\end{tabular}
\label{tab:main}
\end{table*}

\begin{table*}
\caption{Multiple metrics results on real-world multivariate datasets. The P and R are the precision and recall. The $F1_{PA}$ is the F1 score using the PA strategy. Aff-P and Aff-P are the precision/recall pair of affiliation metrics. R\_A\_R and R\_A\_P are Range-AUC-ROC and Range-AUC-PR. V\_ROC and V\_PR are volumes under the surfaces created based on ROC and PR curves, respectively. The best ones are in \textbf{Bold}. }\vspace{-4mm}
\begin{tabular}{lcccccccccccc}
\hline \hline
& Methods & $F1$  &P & R & $F1_{PA\%K}$   & $F1_{PA}$ & Aff-P &  Aff-R & R\_A\_R &  R\_A\_P  & V\_ROC & V\_PR  \\ \hline
\multirow{3}{*}{\textit{\begin{tabular}[c]{@{}l@{}}EMG \end{tabular}}}     
& WETAS      & 0.2446 & \textbf{0.6334} & 0.1516 & 0.3246   & 0.4612 & 0.8498  & 0.2962 & 0.5483 & 0.5338 & 0.5493 & 0.5291    \\
& TreeMIL    & 0.1696 & 0.5689 & 0.0997 & 0.2440   & 0.5688 & 0.7829  & 0.4956 & 0.5863 & 0.5825 & 0.5851 & 0.5764    \\
& \textbf{\PUdetector} & \textbf{0.4431} & 0.5660 & \textbf{0.3640} & \textbf{0.5174}   & \textbf{0.6188} & \textbf{0.8916}  & \textbf{0.5705} & \textbf{0.6579} & \textbf{0.5894} & \textbf{ 0.6556} & \textbf{0.5808}    \\
\hline
\multirow{3}{*}{\textit{\begin{tabular}[c]{@{}l@{}}SMD \end{tabular}}}     
& WETAS      & 0.1020 & 0.0762 & 0.1542 & 0.1171  & 0.1370 & 0.5259  & 0.6530 & 0.5120 & 0.1602 & 0.5115 & 0.1581    \\
& TreeMIL    & 0.0999 & 0.0612 & \textbf{0.2723} & 0.1164  & 0.1601 & 0.5212  & \textbf{0.7880} & 0.5510 & 0.2334 & 0.5477 & 0.2300    \\
& \textbf{\PUdetector} & \textbf{0.1092} & \textbf{0.2232} & 0.0722 & \textbf{0.1630}  &\textbf{ 0.6761} &\textbf{ 0.6363}  & 0.5317 & \textbf{0.6108} & \textbf{0.5029} & \textbf{0.6118} & \textbf{0.5041}    \\
\hline
\multirow{3}{*}{\textit{\begin{tabular}[c]{@{}l@{}}PSM \end{tabular}}}     
& WETAS      & 0.3927 & 0.4651 & 0.3398 & 0.4891  & 0.6532 & 0.6145  & 0.6550 & 0.7128 & 0.7681 & 0.7055 & 0.7549    \\
& TreeMIL    & 0.3494 & 0.4103 & 0.3043 & 0.4208  & 0.6686 & 0.6756  & \textbf{0.7036} & 0.7273 & 0.7799 & \textbf{0.7367} & 0.7766    \\
& \textbf{\PUdetector} & \textbf{0.4990}& \textbf{0.6976} & \textbf{0.3884} & \textbf{0.6356}  & \textbf{0.8856} &\textbf{0.7491} & 0.5374  & \textbf{0.7462} & \textbf{0.7982} & 0.7261 &\textbf{0.7837}    \\
\hline \hline
\end{tabular}
\label{tab:multir}
\end{table*}

\begin{table*}[]
\caption{Performances of \PUdetector and weakly supervised learning baselines on real-world multivariate datasets under different label noise rates. Label noise Rate is the proportion of positive segments seen as unlabeled, $e_1:=P ( \tilde{Y}=0|Y=1 )$. $F1$ is the F1-score, $F1_{PA\%K}$ is the optimized PA-based F1 score. The best ones are in \textbf{Bold}, and the second ones are \underline{underlined}.}\vspace{-3mm}
\centering
\begin{tabular}{c|c|cc|cc|cc|cc|cc}
\hline\hline
& \textbf{Dataset} & \multicolumn{2}{c|}{\textbf{EMG}} & \multicolumn{2}{c|}{\textbf{SMD}} & \multicolumn{2}{c|}{\textbf{PSM}} & \multicolumn{2}{c|}{\textbf{MSL}} & \multicolumn{2}{c}{\textbf{SMAP}} \\ 
\hline
\textbf{Label Noise Rate} & \textbf{Methods} & \multicolumn{1}{c}{\textbf{$F1$}} & \multicolumn{1}{c|}{\textbf{$F1_{PA\%K}$}} & \multicolumn{1}{c}{\textbf{$F1$}} & \multicolumn{1}{c|}{\textbf{$F1_{PA\%K}$}} & \multicolumn{1}{c}{\textbf{$F1$}} & \multicolumn{1}{c|}{\textbf{$F1_{PA\%K}$}} & \multicolumn{1}{c}{\textbf{$F1$}} & \multicolumn{1}{c|}{\textbf{$F1_{PA\%K}$}} & \multicolumn{1}{c}{\textbf{$F1$}} & \multicolumn{1}{c}{\textbf{$F1_{PA\%K}$}} \\ 
\hline
\multirow{4}{*}{\textit{\begin{tabular}[c]{@{}c@{}}0.4\end{tabular}}}  
& DeepMIL &0.0484 & 0.0821 & 0.0865 & 0.1203 & 0.2274 & 0.3374 & 0.0225 & 0.0398 & 0.1391 & 0.1898 \\ 
& WETAS & \underline{0.3906} & \underline{0.4929} & \textbf{0.1393} & \underline{0.1603} & 0.2978 & 0.3884 & 0.0753 & 0.1186 & \underline{0.1506} & \underline{0.2002} \\ 
& TreeMIL & 0.2561 & 0.3437 & 0.1247 & 0.1556 & \underline{0.4665} & \underline{0.5379} & \underline{0.1257} & \underline{0.1494} & 0.1220 & 0.1294 \\ 

& \textbf{\PUdetector} & \textbf{0.5599} & \textbf{0.6212} & \underline{0.1280} & \textbf{0.1823} & \textbf{0.5548} & \textbf{0.6883} & \textbf{0.1395} & \textbf{0.1531} & \textbf{0.1957} & \textbf{0.2012} \\ 
\hline
\multirow{4}{*}{\textit{\begin{tabular}[c]{@{}c@{}}0.2\end{tabular}}}  
& DeepMIL & 0.0924 & 0.1511 & 0.1026 & 0.1306 & 0.1654 & 0.2564 & 0.0145 & 0.0247 & 0.0686 & 0.1149 \\ 
& WETAS & \underline{0.5543} & \underline{0.6530} & \underline{0.1297} & \underline{0.1671} & 0.2110 & 0.3172 & 0.0987 & 0.1509 & 0.0971 & 0.1560 \\ 
& TreeMIL & 0.5284 & 0.6010 & 0.1279 & 0.1510 & \underline{0.3812} & \underline{0.4367} & \underline{0.1126} & \underline{0.1637} & \underline{0.1715} & \textbf{0.2017} \\ 

& \textbf{\PUdetector} & \textbf{0.6057} & \textbf{0.6638} & \textbf{0.1313} & \textbf{0.1810} & \textbf{0.5202} & \textbf{0.5301} & \textbf{0.1745} & \textbf{0.1829} & \textbf{0.1879} & \underline{0.1965} \\ 
\hline

\multirow{4}{*}{\textit{\begin{tabular}[c]{@{}c@{}}0.0\end{tabular}}}  
& DeepMIL & 0.0901 & 0.1446 & 0.0636 & 0.0942 & 0.4276 & 0.5421 & 0.0022  & 0.0039  & 0.1033  & 0.1324 \\ 
& WETAS & \textbf{0.6405} & \textbf{0.7403} & \underline{0.1308} & \underline{0.1715} & 0.4879 & 0.5621 & 0.1089 & \textbf{0.1608} &  0.1381 & 0.1496  \\ 
& TreeMIL & 0.5443 & 0.6040 &  \textbf{0.1769} & \textbf{0.2161} &\textbf{0.5759} & \textbf{0.6229} & \textbf{0.1221}  & \underline{0.1596} & \underline{0.1508} & \textbf{0.1956} \\ 

& \textbf{\PUdetector} & \underline{0.6141} & \underline{0.7207} & 0.1255 & 0.1687 & \underline{0.5130} & \underline{0.5905} &  \underline{0.1144} &  0.1246 & \textbf{0.1704} & \underline{0.1856}  \\ 

\hline\hline
\end{tabular}
\label{tab:nr}
\end{table*}

\textbf{Baselines.} To evaluate the performance of the proposed method, we compare our approach with 13 different methods. Additionally, we conduct PU learning of point-wise anomalies with noisy point labels as a comparison in Appendix~\ref{study_comparison}.
\begin{itemize}[leftmargin=*]
    \item \textbf{Unsupervised Learning.} Due to the popularity of utilizing self-supervised methods to detect temporal anomalies, we compare \PUdetector with DCdetector~\cite{yang2023dcdetector}, Anomaly Transformer~\cite{xu2021anomaly}, and some reconstruction-based methods which compute the point-level anomaly scores from the reconstruction of time series, like AutoFormer~\cite{wu2021autoformer}, FEDformer~\cite{zhou2022fedformer}, TimesNet~\cite{wu2022timesnet}, and One-fits-all~\cite{zhou2023one}.
    \item \textbf{Semi-supervised learning.} The variants of the unsupervised methods, like AutoFormer++, FEDformer++, TimesNet++, and One-fits-all++, except for Anomaly Trasfomer and DCdetector. These models are trained by using only normal segments in order to make them utilize given segment-level labels. But note that the 'normal segments' are actually the unlabeled segments in our setting. Our method is designed in noisy label setting. We aim to compare the noise tolerance of these methods.
    \item \textbf{Weakly supervised learning.} The Multiple Instance Learning (MIL) method, like DeepMIL~\cite{sultani2018real}, encourages high anomaly scores at individual time steps based on the segment-level labels. 
    Also, we compare our method with WETAS~\cite{lee2021weakly} and TreeMIL~\cite{liu2024treemil}, which are the main baselines we need to compare.
   
\end{itemize}

\subsection{Experimental Setting}

\textbf{Evaluation metrics.} We adopt various evaluation metrics for comprehensive comparison, including the commonly used F1 score (denoted by $F1$) using both the segment-level ($F1$-W) and point-level ($F1$-D) ground truth. Unless otherwise specified, all results are point-level.
Note that we do not consider the point adjustment (PA) approach~\cite{shen2020timeseries,xu2021anomaly,yang2023dcdetector} for the evaluation of all methods in the main table.
~\cite{kim2022towards} indicates that PA overestimates classifier performance, even though this metric has practical justifications~\cite{xu2018unsupervised}. 
Thus, we adopt the optimized PA-based metric, PA\%K~\cite{kim2022towards}, which actually calculates the AUC of PA\%K to reduce dependence on the parameter K. Also, we report the recently proposed evaluation measures: affiliation precision/recall pair~\cite{huet2022local} and Volume under the surface (VUS)~\cite{paparrizos2022volume}. Different metrics provide different views for anomaly evaluations.

\textbf{Implementation Details.} 
Following the pre-processing methods in~\cite{xu2021anomaly}, we split the dataset into consecutive non-overlapping segments by sliding window. In the implementation phase, running a sliding window in time series data is widely used in TSAD tasks~\cite{shen2020timeseries} and has little influence on the main design~\cite{yang2023dcdetector}. Thus, we just set the window size L as 100 for all datasets, unlike TreeMIL~\cite{liu2024treemil} and WETAS~\cite{lee2021weakly}. 
We use a simple 6 MLP layers as the backbone of the classifier and use ReLU~\cite{nair2010rectified} activation between every fully connected layer and Sigmoid activation for the last layer respectively. We use the Adam~\cite{kingma2014adam} optimizer with an initial learning rate 0.0001. We set the batch size as 32 and use the batch-size training. In each batch, we identify set of variables on which our loss depends, compute gradient for each variable, and obtain the final gradient through chain rule.
The parameters of smoothness and separability constraints in the time constraint loss are set to $\lambda_1 = \lambda_2 = 8\times10^{-5}$ for the best performance.
The baselines are implemented based on the suggested hyperparameters reported in the corresponding previous literature. 

\subsection{Performances on Anomaly Detection}

We first evaluate our \PUdetector with 13 competitive baselines on five real-world multivariate datasets under both pure $F1$ and $F1_{PA\%K}$ metrics as shown in Table~\ref{tab:main}. It can be seen that our proposed \PUdetector achieves the best results under the pure F1 score and PA\%K F1 score on all benchmark datasets. The unsupervised and semi-supervised methods perform considerably worse than the weakly supervised methods in both $F1$ and $F1_{PA\%K}$. 

For all unsupervised methods, they cannot achieve as high detection performance as the weakly supervised methods, which indicates the importance of label information. 
Semi-supervised techniques, which only use the unlabeled segments (know as much label information as our method) as the training data, marginally outperform unsupervised methods but still lag behind weakly supervised ones. The presence of anomalous segments within unlabeled data highlights a limitation of semi-supervised methods—they struggle to effectively handle noise within `normal' (actually unlabeled) data. This shed light on the importance of how to utilize the positive segment-level labels and tackle the noise in the label information, as these factors significantly influence the final detection performance.

As the recent weakly supervised methods WETAS and TreeMIL achieve better results than other baseline models, we mainly compare \PUdetector with them under different evaluation metrics as shown in Table~\ref{tab:multir}. We can learn from the results that our method performs better than WETAS and TreeMIL in most metrics. Our proposed approach can effectively handle noise, making our method more robust to segment-level label noise, while WETAS and TreeMIL rely more on accurate label information, especially the normal label.  

From Table~\ref{tab:main} and Table~\ref{tab:nr}, We can see our method achieves robust results under different label noise rates. While the other methods are less tolerant of noisy label and show unstable performance under the high label noise rate.
The results demonstrate the limitations of current unsupervised methods, especially when the anomalies lie in the training data for building the normal patterns. When there is less noise in the label, \PUdetector still outperforms WEATS and TreeMIL, but the gap is not very large. For example, when the label noise rate is 0.2, the F1 scores on the EMG dataset only decrease by 0.051 and 0.077, respectively. However, in the label noise rate of 0.6 cases, the \PUdetector completely outperforms the three methods by at least 0.2 on the EMG dataset. When the label noise rate is 0.0, turning the problem into a MIL problem, our method achieved the second-best performance on most datasets, comparable to WETAS and TreeMIL.

\begin{table*}[]
\caption{Ablation Studies on Sample Selector in \PUdetector. 'Extraction' is the process of obtaining reliable negatives (RN) based on the confidence, and 'Propagation' is the process of propagating the label to the neighboring nodes. F1-W denotes the F1 score between the actual segment-level labels and the segment-level predictions. F1-D denotes the F1 score between dense labels and the point-level predictions. F1-D (w/o HOC) denotes the F1 score with random noise rate k.  The best ones are in Bold.}
\centering
\vspace{-3mm}
\scalebox{0.93}{
\begin{tabular}{cc|ccc|ccc|ccc}
\hline\hline
\multicolumn{2}{c|}{\textbf{Sample Selector}} & \multicolumn{3}{c|}{\textbf{EMG}} & \multicolumn{3}{c|}{\textbf{PSM}} & \multicolumn{3}{c}{\textbf{SMAP}} \\ 
\hline
\textbf{Extraction} & \textbf{Propagation} & \multicolumn{1}{c}{\textbf{F1-W}} & \multicolumn{1}{c}{\textbf{F1-D(w/o HOC)}} & \multicolumn{1}{c|}{\textbf{F1-D}} & \multicolumn{1}{c}{\textbf{F1-W}} & \multicolumn{1}{c}{\textbf{F1-D(w/o HOC)}} & \multicolumn{1}{c|}{\textbf{F1-D}} & \multicolumn{1}{c}{\textbf{F1-W}} & \multicolumn{1}{c}{\textbf{F1-D(w/o HOC)}} & \multicolumn{1}{c}{\textbf{F1-D}} \\ 
\hline
\ding{55} & \ding{55} & 0.4643 & 0.3619 & 0.4130 & 0.4875 & 0.3446 & 0.4916 & 0.2256 &0.0917 & 0.1878 \\
\ding{52} & \ding{55} & 0.4294 & 0.3485 & 0.4055 & 0.4731 & 0.4382 & 0.4501 & 0.1417 & 0.0516& 0.1318 \\
\ding{55} & \ding{52} & 0.4125 & 0.3124 &  0.4084  & 0.4792 & \textbf{0.4574} & 0.4676 & 0.2418 & 0.0752 & 0.2009 \\
\ding{52} & \ding{52} & \textbf{0.4824} & \textbf{0.3800} & \textbf{0.4431} & \textbf{0.4920} & 0.4566 & \textbf{0.4990} & \textbf{0.2449} &\textbf{0.2097} & \textbf{0.2367} \\
\hline\hline
\end{tabular}
}
\label{tab:ab-us}
\end{table*}

\begin{table*}[]
\caption{Ablation Studies on PU Criterion in \PUdetector. F1-W denotes the F1 score between actual segment-level labels and the segment-level predictions. F1-D denotes the F1 score between dense labels and the point-level predictions. F1-D (w/o HOC) denotes the F1 score with random noise rate k.  The best ones are in Bold.}\vspace{-3mm}
\centering
\scalebox{0.93}{
\begin{tabular}{ccc|ccc|ccc|ccc}
\hline\hline
\multicolumn{3}{c|}{\textbf{PU Criterion}} & \multicolumn{3}{c|}{\textbf{EMG}} & \multicolumn{3}{c|}{\textbf{PSM}} & \multicolumn{3}{c}{\textbf{SMAP}} \\ 
\hline
\textbf{$L_{bce}$} & \textbf{$R_{pu}$} & \textbf{$L_{tc}$} & \multicolumn{1}{c}{\textbf{F1-W}} & \multicolumn{1}{c}{\textbf{F1-D(w/o HOC)}} & \multicolumn{1}{c|}{\textbf{F1-D}} & \multicolumn{1}{c}{\textbf{F1-W}} & \multicolumn{1}{c}{\textbf{F1-D(w/o HOC)}} & \multicolumn{1}{c|}{\textbf{F1-D}} & \multicolumn{1}{c}{\textbf{F1-W}} & \multicolumn{1}{c}{\textbf{F1-D(w/o HOC)}} & \multicolumn{1}{c}{\textbf{F1-D}} \\ 
\hline
\ding{52} & \ding{55} & \ding{55} & 0.3871 & 0.3110 & 0.3695 & 0.4712 & 0.3159 & 0.4811 & 0.2192 & 0.1772 & 0.1794 \\
\ding{52} & \ding{55} & \ding{52} & 0.4294 & 0.3485 & 0.4055 & 0.4792 & 0.3674 & 0.4648 & 0.2071 & 0.0793 & 0.1713 \\
\ding{55} & \ding{52} & \ding{55} & 0.4734 & 0.3751 & 0.4406 & 0.4876 & 0.3446 & 0.4802 & 0.1743 & 0.0971 & 0.1534 \\
\ding{55} & \ding{52} & \ding{52} & \textbf{0.4824} & \textbf{0.3800} & \textbf{0.4431} & \textbf{0.4920} & \textbf{0.4566} & \textbf{0.4990} & \textbf{0.2449} &\textbf{ 0.2097} & \textbf{0.2367} \\
\hline\hline
\end{tabular}
}
\label{tab:ab-pu}
\end{table*}

\begin{table*}[]
\caption{Performances of \PUdetector and point-level PU learning method on real-world multivariate datasets when the segment-level label noise rate is 0.6. $F1$ is the F1-score, $F1_{PA\%K}$ is the optimized PA-based F1 score. The best ones are in \textbf{Bold}.}
\centering
\vspace{-3mm}
\begin{tabular}{c|cc|cc|cc|cc|cc}
\hline\hline
\textbf{Dataset} & \multicolumn{2}{c|}{\textbf{EMG}} & \multicolumn{2}{c|}{\textbf{SMD}} & \multicolumn{2}{c|}{\textbf{PSM}} & \multicolumn{2}{c|}{\textbf{MSL}} & \multicolumn{2}{c}{\textbf{SMAP}} \\ 
\hline
\textbf{Metric} & \multicolumn{1}{c}{\textbf{$F1$}} & \multicolumn{1}{c|}{\textbf{$F1_{PA\%K}$}} & \multicolumn{1}{c}{\textbf{$F1$}} & \multicolumn{1}{c|}{\textbf{$F1_{PA\%K}$}} & \multicolumn{1}{c}{\textbf{$F1$}} & \multicolumn{1}{c|}{\textbf{$F1_{PA\%K}$}} & \multicolumn{1}{c}{\textbf{$F1$}} & \multicolumn{1}{c|}{\textbf{$F1_{PA\%K}$}} & \multicolumn{1}{c}{\textbf{$F1$}} & \multicolumn{1}{c}{\textbf{$F1_{PA\%K}$}} \\ 
\hline
Point-level nnPU~\cite{kiryo2017positive} & 0.1077 & 0.1077 & 0.0735 & 0.0735 & 0.4538 & 0.4538 & 0.0775 & 0.0775  & 0.1293 & 0.1293 \\ 
Point-level Dist-PU~\cite{zhao2022dist} & 0.2944 & 0.3726 & 0.0735 & 0.0735 & 0.4451 & 0.5799 & 0.0983 &  0.1498 & 0.1223 & 0.1743 \\ 
Point-level NCAD~\cite{carmona2021neural} & 0.0651 & 0.1563 & 0.0676 & \textbf{0.9318} & 0.0159 & 0.0182 & 0.0414 & 0.0987 & 0.0217 & 0.0254 \\
Segment-level \PUdetector (Ours) & \textbf{0.4431} & \textbf{0.5174} & \textbf{0.1092} & 0.1630 & \textbf{0.4990} & \textbf{0.6356} & \textbf{0.2029} & \textbf{0.2219} & \textbf{0.2367} & \textbf{0.2906} \\ 
\hline\hline
\end{tabular}
\label{tab:point-level comparision}
\end{table*}

\subsection{Ablation Studies}

To better understand how each component in \PUdetector contributes to the final accurate anomaly detection, we conduct extensive ablation studies. 

\textbf{Model Structure.} In the temporal embedding module, WETAS is adopted as a representation framework to get the informative temporal features. We conduct an ablation study on how the representation model structure (both Di-CNN and Transformer~\cite{vaswani2017attention} are seven layers) affects the performance of our method in Table~\ref{tab:ab-m} in the Appendix. The results indicate that Di-CNN and Transformer architectures have basically the same performance, but the former performs slightly better in multiple metrics.

\textbf{Sample Selector.} With confidence-based sample selection modules, we can see that \PUdetector gains the best performance from Table~\ref{tab:ab-us}. Both confidence-based extraction and label propagation processes will degrade the performance of our model when used individually. However, if they are utilized simultaneously, the performance improves. This may be because only reliable extraction loses most of the data information, while using only label propagation may lead to more pseudo-positive segments in the similarity-based network. 

\textbf{PU Criterion.} Table~\ref{tab:ab-pu} shows the ablation study of PU Criterion. According to the loss function definition in Section~\ref{sec:criterion}, we use PU Loss, denoted as $R_{pu}$, and TC Loss, denoted as $L_{tc}$. To utilize the provided segment-level labels, we use the binary cross entropy as the basic loss function. We can find that our PU Criterion leads to a great improvement from those without PU Loss. Also, the results demonstrate the effectiveness of the proposed time constraint loss. 

We find that HOC~\cite{zhu2021clusterability} can automate the threshold selection process, which improves the performance of point-level detection. These ablation studies demonstrate the effectiveness of our approach.

\subsection{Additional Analysis}
We aim to study the flexibility provided by our method, which utilizes only partial segment-level labels, and compare its detection performance with that of point-level PU learning methods and augmentation-based methods (see Appendix~\ref{study_comparison}).

\textbf{Point-level PU learning.}
We use the same temporal embedding as our method to represent the temporal feature of each time series point for a fair comparison, as these methods~\cite{kiryo2017positive, zhao2022dist} are not specifically designed for time series data. 
In terms of input data, these methods require point-level positive data and unlabeled data. For all labeled anomaly segments in our setting (segment-level label noise rate = 0.6), the anomaly points within these segments constitute the positive dataset, while the normal points within these segments and points in other segments (both actual anomaly segments and normal segments) constitute the unlabeled dataset. This ensures that the label information known to the point-level PU method is as close as possible to ours, avoiding the situation where knowing too many labels renders the comparison meaningless.

Since NCAD~\cite{carmona2021neural} is not the most recent SOTA unsupervised learning method, we use the provided supervised setting. However, because point-level labels are required, we provide point-level labels  based on the segment-level label noise rate of 0.6 (as mentioned above). Note that this kind of labeling actually contains more information than segment-level labeling.

The results, shown in Table~\ref{tab:point-level comparision}, indicate that our approach performs better than point-level PU learning methods while providing greater flexibility by utilizing only segment-level labels. It also outperforms NCAD using noisy point-level labels.

\subsection{Hyperparameter Studies}
\begin{figure}[!hbtp]
\centering
\includegraphics[width=0.48\textwidth]{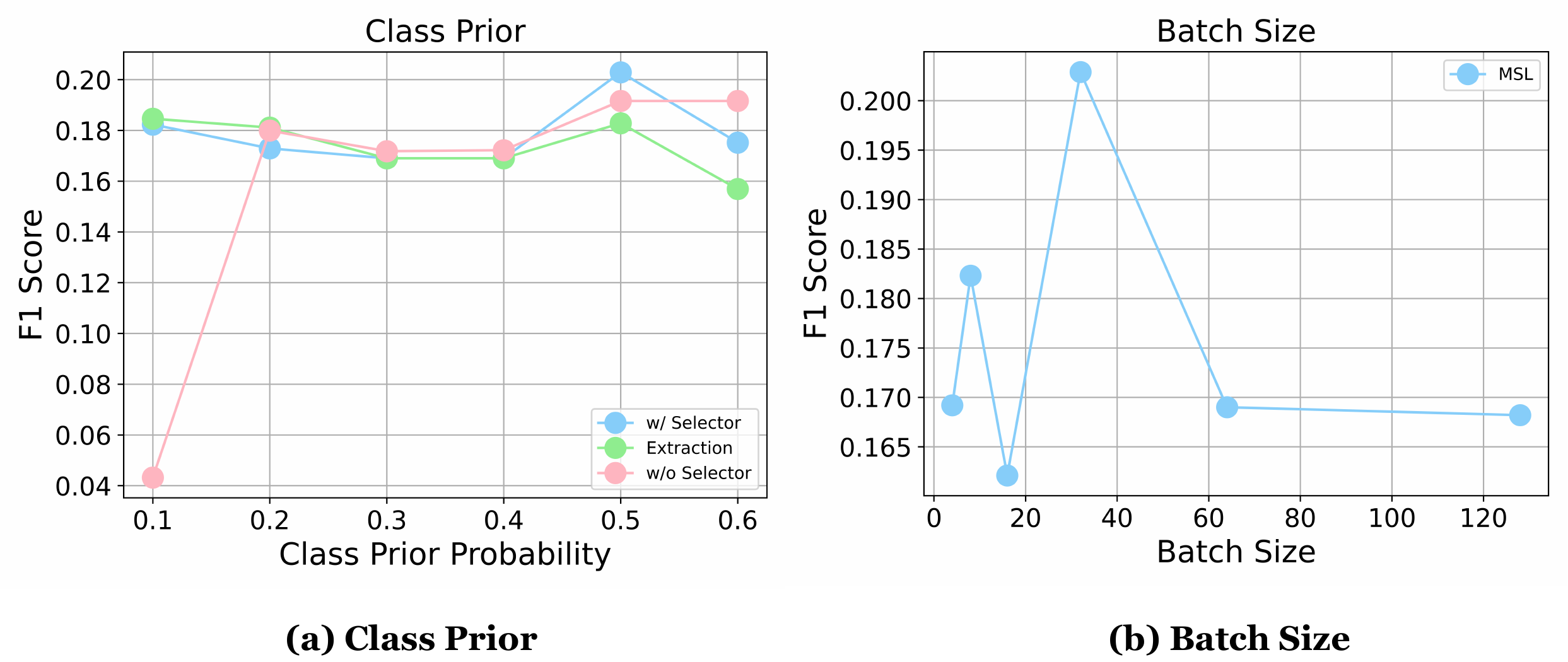}
\caption{Parameter sensitivity studies of hyper-parameters, Class Prior and Batch Size in NRdetector. Both studies are conducted on the MSL dataset.} 

\label{fig:param}
\end{figure}

To evaluate the robustness of our framework under different hyperparameters, we vary the parameters to investigate how the model performance varies.
Figure~\ref{fig:param}(a) shows the performance on the MSL dataset under different Class Prior. 
The result demonstrates that \PUdetector is robust with a wide range of class prior, alleviating the need for PU Loss to estimate class prior accurately. Figure~\ref{fig:param}(b) shows the performance under different batch sizes. It indicates that \PUdetector achieves the best performance with the batch size 32. More experimental results are left in Appendix~\ref{app:exp}.

\section{Conclusion}
We have proposed NRdetector, a noise-resilient multivariate TSAD framework with a novel contrastive-based loss function to bridge the gap between noisy segment labels and missing point labels. 
We have experimentally demonstrated the effectiveness of each component in our framework. Comparisons with state-of-the-art baselines on several real-world time series benchmarks reveal that \PUdetector consistently outperforms existing time series anomaly detection methods in the noisy label setting.
Future directions involve exploring the power of large language models (LLMs) on such tasks and implementing this approach in real-world label noise scenarios.

\clearpage
\balance
\bibliographystyle{ACM-Reference-Format}
\bibliography{sample-base}

\appendix

\section{Algorithms}
\label{app:algorithms}

\IncMargin{1em} 
\begin{algorithm}[!h]
    \SetAlgoNoLine
    \SetKwInOut{Input}{\textbf{Input}}\SetKwInOut{Output}{\textbf{Output}} 
    \Input{
        \\
        $\mathcal X_L$, a set of labeled positive segments\\
        $\mathcal X_U$, a set of unlabeled segments\\
        $m$, number of iterations\\
        $\lambda_0$, controls the size of set to be filtered out\\
        $W$, similarity matrix calculated by Katz index}
    \Output{
        \\
        $\mathcal X_{RN}$, set of extracted reliable positive segments\\}
    \BlankLine

    initialize the $\mathcal X_{RN}$ and $\mathcal X_{out}$ set\; 
    initialize k = 0\;
    \Repeat
        {\text{k=$m$}}
        {based on $W$, calculate $S_{X_i}$, $S_{X_i}=\frac{\sum_{j=1}^{N_P}{W_{i,j}}}{N_p}$; // \textcolor{blue}{\emph{$S_{X_i}$ is the mean similarity between an unlabeled segment $X_i$ and all labeled positive segments}}\;
        rank each segment $X_i$ according to the $S_{X_i}$, $X_i\in \mathcal X_U$\;
        $out'$ $\gets$ the top $\frac{\lambda _0}{m}\times N_P$ ranked examples in $\mathcal X_U$\;
        $\mathcal X_{out}\gets \mathcal X_{out}\cup \mathcal X_{out'}$, $\mathcal X_U\gets \mathcal X_U-\mathcal X_{out'}$\;
        $k=k+1$\;
        
        }
        based on $W$, calculate $S_{X_i}$, $X_i\in  \mathcal X_U-\mathcal X_{out}$, $S_{X_i}=\frac{\sum_{j=1}^{N_P+|out|}{W_{i,j}}}{N_p+|out|}$\;
        rank each segment $X_i$ according to the  $S_{X_i}$, $X_i\in \mathcal X_U-\mathcal X_{out}$\;
        $\mathcal X_{RN}$ $\gets$ the bottom $N_P+|\mathcal X_{out}|$ ranked examples in $ \mathcal X_U-\mathcal X_{out}$ \;
        return $\mathcal X_{RN}$
    \caption{Sample Selector: Confidence-based Extraction \label{al1}}
\end{algorithm}

\DecMargin{1em}
\IncMargin{1em} 
\begin{algorithm}[!h]
    \SetAlgoNoLine 
    \SetKwInOut{Input}{\textbf{Input}}\SetKwInOut{Output}{\textbf{Output}}
    \Input{
        \\
        $X$, a set of all segments\\
        $F_X$, the label information 
        \textcolor{blue}{\emph{//For $X_i\in \mathcal X_P$, the label is [0,1];for $X_i \in \mathcal X_{RN}$, the label is [1,0]; the label of other segments are [0,0]}} 
        
        \\
        $w$, connection weights between nodes}
    \Output{
        \\
        $F_{\mathcal X_U}$, class information for unlabeled segments\\}
    \BlankLine

    $D \gets diag\left( w\cdot I_N\right)$; \textcolor{blue}{\emph{// diag(...) is the matrix diagonal operator}}\;
    $P\gets \left( D^{-1} \right) \cdot w$ \;
    \textbf{while} there is no convergence or maximum number of iterations not being reached \textbf{do}\\
        $F_X \gets P \cdot F_X$ \;
        $F_{\mathcal X_L} \gets Y_{\mathcal X_L}$ \;
    end \textbf{while} \\
    return $F_{\mathcal X_U }$
       
    \caption{Sample Selector: Label Propagation \label{al2}}
\end{algorithm}
\DecMargin{1em}

\begin{table*}[]
\caption{Ablation Studies on model structure for temporal embedding module in NRdetector. F1-W denotes the F1 score between the actual segment-level labels and the segment-level predictions. F1-D denotes the F1 score between dense labels and the point-level predictions. F1-D (w/o HOC) denotes the F1 score with random anomaly rate k. The best ones are in Bold.}
\centering
\vspace{-3mm}
\begin{tabular}{c|ccc|ccc|ccc}
\hline\hline
\textbf{Dataset} & \multicolumn{3}{c|}{\textbf{EMG}} & \multicolumn{3}{c|}{\textbf{PSM}} & \multicolumn{3}{c}{\textbf{SMAP}} \\ 
\hline
\textbf{Model Structure} & \multicolumn{1}{c}{\textbf{F1-W}} & \multicolumn{1}{c}{\textbf{F1-D(w/o HOC)}} & \multicolumn{1}{c|}{\textbf{F1-D}} & \multicolumn{1}{c}{\textbf{F1-W}} & \multicolumn{1}{c}{\textbf{F1-D(w/o HOC)}} & \multicolumn{1}{c|}{\textbf{F1-D}} & \multicolumn{1}{c}{\textbf{F1-W}} & \multicolumn{1}{c}{\textbf{F1-D(w/o HOC)}} & \multicolumn{1}{c}{\textbf{F1-D}} \\ 
\hline
Di-CNN & \textbf{0.4824} & \textbf{0.3800} & \textbf{0.4431} & 0.4780 & \textbf{0.3647} & \textbf{0.4990} & \textbf{0.2449} & \textbf{0.2097} & \textbf{0.2367} \\
Transformer & 0.4734 & 0.3751 & 0.4406 & \textbf{0.5301} & 0.3444 & 0.4930 & 0.2410 & 0.2051 & 0.2354 \\
\hline\hline
\end{tabular}\vspace{-2mm}
\label{tab:ab-m}
\end{table*}

\begin{table*}[]
\caption{Experimental results comparing our method with CARLA and CutAddPaste on the EMG, SMD, and PSM datasets under a 0.6 label noise rate. F1-W denotes the F1 score between the actual segment-level labels and the segment-level predictions.}
\centering
\vspace{-3mm}
\scalebox{0.93}{
\begin{tabular}{c|ccc|ccc|ccc}
\hline\hline
 & \multicolumn{3}{c|}{\textbf{EMG}} & \multicolumn{3}{c|}{\textbf{SMD}} & \multicolumn{3}{c}{\textbf{PSM}} \\ 
\hline
\textbf{Method} & \multicolumn{1}{c}{\textbf{F1-W}} & \multicolumn{1}{c}{\textbf{$F1$}} & \multicolumn{1}{c|}{\textbf{$F1_{PA\%K}$}} & \multicolumn{1}{c}{\textbf{F1-W}} & \multicolumn{1}{c}{\textbf{$F1$}} & \multicolumn{1}{c|}{\textbf{$F1_{PA\%K}$}}  & \multicolumn{1}{c}{\textbf{F1-W}} & \multicolumn{1}{c}{\textbf{$F1$}} & \multicolumn{1}{c}{\textbf{$F1_{PA\%K}$}}  \\ 
\hline
CARLA~\cite{darban2025carla} & 0.1918 & 0.0939 & 0.1044 & 0.1383 & 0.0642 & 0.0793  & 0.5589 & 0.1882 & 0.1882 \\
CutAddPaste(floating)~\cite{wang2024cutaddpaste} &0.4153 & 0.1869 & 0.2362 & 0.1754 & 0.0699 & 0.0844  & 0.5589 & 0.1882 & 0.1882  \\
\PUdetector (Ours) & 0.4824 & 0.4431 & 0.5174 & 0.1693 & 0.1092 & 0.1630 & 0.4780 & 0.4990 & 0.6356 \\
\hline\hline
\end{tabular}
}
\label{tab:ab-point-aug}
\end{table*}
\section{Proof for Theorems}
\label{app:theo}

\subsection{Upper Bound}
\label{app:theo-bound}

We consider the case where $Y|X$ is confident, i.e., each feature $X$ belongs to one particular true class $Y$ with probability 1, which is generally held in classification problems~\cite{liu2015classification}. Due to the specificity of the PU setting, where $e_0$ equals 0, we could get the following theorem based on Lemma 2 and Theorem 2 in~\cite{zhu2021rich}. 

\begin{theorem}[]\label{thm:noisy_bound}
With probability at least $1-\delta$, the generalization error on datasets $\tilde{D}$ is upper-bounded by
\[
R_{\mathcal D}( \hat{f} ) \le \bar\eta+\sqrt{\frac{2\log \left( \frac{4}{\delta} \right)}{N}}+P( \tilde{Y}\ne \tilde{Y}^* ), 
\]
where $\bar\eta$ is the expected label error in the pseudo noisy dataset $\tilde{D}$.
\end{theorem}

\begin{lemma}[]\label{lem:eta_error}
For any feature $X$, if $Y|X$ is confident, $\eta(X)$ is the error rate of the model prediction on $X$, i.e., $\exists i \in[K]: P(Y=i|X)=1 \Rightarrow \eta(X)= P(\tilde Y \ne Y|X)=e(X).$
\end{lemma}

Based on Lemma~\ref{lem:eta_error}, for PU setting, we have $\eta(X)= P(\tilde Y \ne Y|X)=e(X)$. In PU setting, $e_0=P( \tilde{Y}=1|Y=0 )=0 $, $e_1:=P( \tilde{Y}=0|Y=1 )$ is the label noise rate. So here we have $\eta(X)=e_1$, and the expected label error is equal to the label noise rate, that is $\bar{\eta}=\int_X{\eta \left( X \right) P\left( X \right) dX}=e_1$. Thus, we could obtain the following theorem based on the Theorem~\ref{thm:noisy_bound}.

\begin{theorem}[Generalization Error]\label{thm:ge}
With probability at least $1-\delta$, the generalization error on datasets $\tilde{D}$ is upper-bounded by
\[
R_{\mathcal D}( \hat{f} ) \le e_1+\sqrt{\frac{2\log \left( \frac{4}{\delta} \right)}{N}}+P( \tilde{Y}\ne \tilde{Y}^* ), 
\]
where $\tilde{D}=\left\{ ( X_n,\tilde{Y}_n ) \right\} _{n\in \left[ N \right]}$ is the noisy dataset, $\hat{f}$ is the classifier trained on the noisy dataset, $e_1$ is the label noise rate, and $\tilde Y^*$ is the prediction of the Bayes optimal classifier $f^*$.
\end{theorem}

As the label noise rate $e_1$ gets smaller, the upper bound for the generalization error of classifier $f$ also decreases when the benefit of label noise rate reduction is more significant than the harm of losing training samples.

\section{More experiments and discussions}
\label{app:exp}

\subsection{Additional Studies}
\label{study_comparison}
\textbf{Augmentation-based methods.} Since our goal is point-level prediction, we adapted augmentation-based methods by assigning anomaly labels to all points within a sample if the sample is predicted as anomalous. We also include segment-level results for reference, even though our focus is point-level prediction using noisy segment-level labels. As shown in Table~\ref{tab:ab-point-aug}, our method outperforms the others in all three metrics for point-level prediction, which is the focus of our paper. We believe that the performance differences arise due to the different target settings of the three methods and do not necessarily indicate which approach is better or worse.

\subsection{Hyperparameters Details}
\begin{table}[t]
\caption{The values of hyperparameters used in \PUdetector}
\begin{tabular}{c|ccccc}
\hline\hline
 & \textbf{EMG} & \textbf{SMD} & \textbf{PSM} & \textbf{MSL} & \textbf{SMAP}\\ 
\hline
$prior$ & 0.25 & 0.8 & 0.4 & 0.5 & 0.5  \\
$anoamly\_ratio$ & 0.65 & 0.15 & 0.6 & 0.8 & 0.9  \\
\hline\hline
\end{tabular}
\label{tab-app:hypeprama}
\end{table}

\PUdetector is implemented in PyTorch, and some important parameter values used in the proposed metrod are listed in Table~\ref{tab-app:hypeprama}. The dimension of the temporal embedding $d\_model$ is 64. In the process of confidence-based sample selection, we use $m$ to control the number of iterations and use $\lambda_0$ to control the size of set to be filtered out. For all datasets, we set $m=4$ and $\lambda_0 =0.32$. $batch\_size$ is set to 32, $learning\_rate$ is set to $1e-5$, and $window\_size$ is 100. $prior$ is the estimation of $\pi_P$ in PU loss (see~\ref{sec:criterion}). $anoamly\_ratio$ is the predefined point-level anomaly rate in the data-centric point-level detection module.

\end{document}